\documentclass{article}

\usepackage{amssymb}\usepackage{amsmath}
\usepackage{graphicx}        % standard LaTeX graphics tool
\usepackage{url}
\usepackage{multirow}
\usepackage[hang,small,bf]{caption}
\newcommand{\ma}[1]{\ensuremath{\mathbf{#1}}}
\newcommand*\xbar[1]{\hbox{\vbox{\hrule height 0.5pt\kern0.5ex\hbox{\kern-0.1em\ensuremath{#1}\kern-0.1em}}}}
\newtheorem{proposition}{Proposition}

\begin{document}
\title{Clustering Binary Data by Application of Combinatorial Optimization Heuristics}
\author{Javier Trejos-Zelaya\thanks{CIMPA \& School of Mathematics, Faculty of Science, University of Costa Rica, San Jos\'e, Costa Rica. E-Mail: javier.trejos@ucr.ac.cr}
\and Luis Eduardo Amaya-Brice\~no\thanks{Guanacaste Campus, University of Costa Rica, Liberia, Costa Rica. E-Mail: luis.amaya@ucr.ac.cr}
\and Alejandra Jim\'enez-Romero\thanks{School of Mathematics, Costa Rica Institute of Technology, Cartago, Costa Rica. E-Mail: alejimenezr@gmail.com}
\and Alex Murillo-Fern\'andez\thanks{Atlantic Campus, University of Costa Rica, Turrialba, Costa Rica. E-Mail: alex.murillo@ucr.ac.cr}
\and Eduardo Piza-Volio\thanks{CIMPA \& School of Mathematics, Faculty of Science, University of Costa Rica, San Jos\'e, Costa Rica. E-Mail: eduardo.piza@ucr.ac.cr}
\and Mario Villalobos-Arias\thanks{CIMPA \& School of Mathematics, Faculty of Science, University of Costa Rica, San Jos\'e, Costa Rica. E-Mail: mario.villalobos@ucr.ac.cr}
}
\maketitle
\begin{abstract}
We study clustering methods for binary data, first defining aggregation criteria that  measure the compactness of clusters. Five new and original methods are introduced, using neighborhoods and population behavior combinatorial optimization metaheuristics: first ones are simulated annealing, threshold accepting and tabu search, and the others are a genetic algorithm and ant colony optimization. 
The methods are implemented, performing the proper calibration of parameters in the case of heuristics, to ensure good results. From a set of 16 data tables generated by a quasi-Monte Carlo experiment, a comparison is performed for one of the aggregations using $L_1$ dissimilarity, with hierarchical clustering, and a version of k-means: partitioning around medoids or PAM. 
Simulated annealing perform very well, especially compared to classical methods.
\end{abstract}

\section{Introduction}
\label{introduction}

Binary data arise in several situations in research since they can encode situations where the presence (1) or absence (0) of a characteristic is studied: species present/absent, alive/dead in health sciences, yes/no in social or decision sciences, and so on. For instance, in Ecology \cite{Salas}, 
it is usual to divide an area in sectors and record the presence or absence of certain vegetable species.
In the study of gene expression data, several molecular techniques encode data as binary matrices \cite{Demey}.
In pattern recognition, images may also be coded as 0/1 data indicating the presence or absence of a feature.
In Sociology \cite{Borkulo}, Health Sciences \cite{Zhang}, Economics \cite{Jeliazkov} \ldots binary data are analyzed.  

Several methods have been used for clustering binary data.
For instance, there are partitioning methods such as dynamical clusters, which is an adaptation of Forgy's k-means \cite{Everitt}, based on a representation of clusters by a kernel and iterations of two steps: allocating objects to the nearest kernel and recalculating the kernels.
A variant of this k-means method is PAM \cite{Kaufman:Roosseuw}, {\it partitioning around medoids}, where kernels are 0/1 median vectors and L1 dissimilarity is used. Another variant is for kernels selected as the object in the class that minimizes the sum of dissimilarities to the rest of objects in the class; this last version is what we will call k-means for 0/1 data in this article. These methods find local minima of the criterion to be minimized, since they are based on local search procedures.

There are also hierarchical methods used for clustering binary data \cite{Everitt}, using an appropriate dissimilarity for binary data (such as Jaccard, for instance). These methods find also local minima since they are based on a greedy strategy.

To overcome the local optima problem, optimization strategies have been used. We have employed combinatorial optimization metaheuristics for clustering numerical data \cite{IFCS:1998}, \cite{IEEE:2014}. In the present article, we will used some of these heuristics for binary data, whenever it is possible. When dealing with binary data it is necessary to adapt the criterion, since Huygens theorem and other theoretical results only hold in an Euclidean context.

The article is organized as follows. Section \ref{clust:binary:data} presents the clustering problem and particularly some criteria and properties for the binary case.
Section \ref{sec:heuristics} contains the five combinatorial optimization metaheuristics employed here and the adaptation we made for clustering binary data.
In Section \ref{sec:results} the results obtained are presented, and finally in Section \ref{sec:conclusion} some concluding remarks are made.

\pagebreak

\section{Clustering binary data}
\label{clust:binary:data}

Given a data set of binary vectors $\Omega = \{\ma{x}_1,\ma{x}_2,\ldots,\ma{x}_n\}$ 
with $\ma{x}_i\in \{0,1\}^p$ and a number $K\in\mathbb{N}$, we seek for a partition
$P = (C_1,C_2,\ldots,C_K)$ of $\Omega$ such that elements in a class $C_k$ are more similar than elements of other classes.

%Criteria
We define a within inertia measure of the partition $P$ as%\vspace{-1mm}
\begin{equation}
W(P) = \sum_{k=1}^K \delta(C_k)
\label{criterion:W}
\end{equation}
where $\delta(C_k)$ can be defined (among others) as% \vspace{-1mm}
\begin{eqnarray*}
\delta_{sum}(C_k) &=&  \sum_{i,i'\in C_k} d(\ma{x}_i,\ma{x}_{i'})
\label{aggreg:WS}\\
\delta_{L_1}(C_k) &=&  \sum_{i\in C_k} \Vert\ma{x}_i-m(C_k)\Vert_{L_1}
\end{eqnarray*}%\vspace{-1mm}
$d$ being a binary dissimilarity index in $\Omega$ (for instance, Jaccard or $L_1$), $m(C_k)$ the median vector of $C_k$.
In \cite{Spath} and \cite{Koln} there are studied some other criteria for clustering.
These indexes satisfy the following \textbf{monotonicity property}. %, that is, the value of the objective function $W(P)$ of the optimization problem's solution for $K+1$ classes is no greater than the corresponding value of the optimization problem for $K$ classes.
 
\begin{proposition}[see \cite{Koln}]
\label{monotonicity}
Let $P=(C_1,\ldots,C_K)$ and $P'=(C_1',\ldots,C_{K+1}')$ be partitions of
$\Omega$ in $K$ and $K+1$ non-empty classes,  $\delta_{sum}$ and $\delta_{L_1}$ on $2^\Omega$ satisfy the monotonicty property, since for all instances of the data, we have
\[ \min_{P'\in{\cal P}_{k+1}^\star} W(P') =
      \sum_{j=1}^{k+1} \delta(C'_j) \le
   \min_{P\in{\cal P}_k^\star} W(P)  =  \sum_{j=1}^k \delta(C_j), \]
for every number of classes $K<n$. 
\end{proposition}

Proofs can be found in \cite{Koln} or by request upon the authors.
As a consequence of proposition \ref{monotonicity} the number of clusters must be predefined, since best clusterings with different number of classes cannot be compared.
In \cite{Koln} it is also proved that for $\delta_{sum}$ and $\delta_{L_1}$ all optimal partitions have non empty classes.

Analogously to the continuous case, total inertia can be defined as:%\vspace{-1mm}
\begin{equation}
I(\Omega) = \sum_{i=1}^n \sum_{i'=1}^n d(\ma{x},\ma{x'})
\label{inertia:I}
\end{equation}%\vspace{-1mm}
and, thanks to the monotonicity, the between-classes inertia is defined as:%\vspace{-1mm}
\begin{equation}
B(\Omega) = I(P) - W(P),
\label{inertia:B}
\end{equation}
and it is always positive.

\section{Using combinatorial optimization heuristics}
\label{sec:heuristics}

We have implemented clustering algorithms using well-known combinatorial optimization heuristics. Three of them are based on neighbors, simlated annealing (SA), threshold accepting (TA) and tabu search (TS), and two based on a population of solutions, genetic algorithm (GA) and ant colony optimization (AC).

A state of the problem is a partition $P$ of $\Omega$ in $K$ clusters.
From a current state, a neighbor is defined by the single transfer of an object from its current class to a new one, chosen according to the corresponding heuristic rules. This will be the case in SA, TA and TS, and the mutation operation in GA.
\vspace{-0.5cm}

\subsubsection*{Simulated annealing}

Simulated annealing is a random search algorithm \cite{Kirkpatrick} that uses an external parameter of control $c_t$ called \textit{temperature}, that controls the random acceptance of states worse than the previous one. It is employed the {\bf Metropolis rule} for this acceptation:
a new state $P'$ generated from $P$ is accepted if $\Delta W < 0$, where $\Delta W = W(P') - W(P)$,  otherwise, it can be accepted with probability $\exp(-\Delta W)/c_t$.
It is well known \cite{Aarts} that under a Markov chain modeling simulated annealing has asymptotic convergence properties to the global optimum under some conditions, so it  is expected that its use permits to avoid local minima. We found some simplification properties for $\Delta W$, calculated in each iteration and useful for speeding up the algorithm.

SA parameters are: $\chi_0$ the initial acceptation rate, $L$ length of Markov chains, $\gamma$ factor reduction for $c_t$ such that $C_{t+1} = \gamma c_t$ and $\epsilon$ stop criterion.
\vspace{-0.3cm}

\subsubsection*{Threshold accepting}

TA was proposed by \cite{Dueck:Scheuer} and can be seen as a particular case of SA, with a different rule for acceptation.
Movements that produce an improvement for the objective function in a neighborhood are accepted
and movements that worsen it are accepted if they fall into a threshold that is positive and decreases in time. Clearly, the acceptation rule is deterministic, not stochastic.

TA parameters are $Th_0$ the initial threshold, $\gamma$ the factor reduction for threshold $Th$, the maximum number of iterations and $\epsilon$ the stop criterion.
\vspace{-0.3cm}

\subsubsection*{Tabu search}

TS \cite{Glover} handles a tabu list $T$ of length $|T|$ with solutions or codes of solutions, that are forbidden to be attained for a certain number of iterations. In each step, current state moves to the best neighbor outside $T$. In our partitioning problem, $T$ stores a code of the transfers that define the neighbors, in this case the indicator function of cluster that contained the object that changed of class during the transfer; that is, all objects that were together in the previous state, are forbidden to be together for $|T|$ iterations.

TS parameters are $|T|$, maximum number of iterations and sampling size of neighborhoods.
\vspace{-0.3cm}

\subsubsection*{Genetic algorithm}

GA handles a population of solutions, which are \textit{chromosomes} representing partitions. A chromosome is an $n$-vector in an alphabet of $K$ numbers indicating the presence/absence of the $i$-th object in the corresponding class. The \textit{fitness} function is defined as $f(P) = \frac{B(P)}{I(\Omega)}$.
In the genetic algorithm iterations, chromosomes are kept using a random wheel roulette with elitism, with a probability proportional to $f$.
For crossover between two parents selected at random (with a uniform distribution), the dominant parent is the one with the greatest fitness; a class in it is selected uniformly at random and this class is copied in the other parent, generating a new son.
Mutation is the classical one: an object is selected randomly, and it is transferred  to a new class.

Parameters for GA are the number $M$ of partitions (population size), probabilities of crossover and mutation, maximum number of iterations. Iterations stop when the fitness variance of the population is less than $\epsilon$.
\vspace{-0.3cm}

\subsubsection*{Ant colonies}

In AC \cite{Dorigo}, each ant $m$ is associated with a partition $P$, which is modified during the iterations.
Given an object $\ma{x}_i$, the probability of transfering another object $\ma{x}_{i'}$ to the same class as $\ma{x}_i$ in iteration $t$ is defined as
\begin{equation}
p_{ii'} = \frac{(\tau_{ii'})^\alpha (\eta_{ii'})^{\beta}}{\sum_{l\neq i}^n (\tau_{li'})^\alpha (\eta_{li'})^{\beta}}
\label{proba}
\end{equation}
where  $\eta_{ii'} = \frac{1}{d_{ii'}}$ is the visibility and the \textbf{pheromone trail} is updated with
\begin{equation}
\tau_{ii'}(t+1) = (1-\rho)\tau_{ii'}(t) + \rho \sum_{m=1}^M (\Delta^m \tau_{ii'}(t+1))
\label{tau}
\end{equation}
\begin{equation}
\Delta^m \tau_{ii'}(t+1) = \left\{ \begin{array}{cl}
\frac{B(P^m)}{I(\Omega)} & \mbox{ if $i,i'\in$ same class}\\
0 & \mbox{otherwise.}\\
\end{array}\right.
\label{Delta}
\end{equation}
$M$ being  the number of ants, $\alpha,\beta\in\mathbb{R}$ weights, and $\rho\in\mathbb{R}$ an evaporation parameter.

AC parameters to calibrate are $\alpha, \beta, \rho$, size of population $M$, precision $\epsilon$ and the maximum number of iterations.

\section{Simulated data}
\label{simulated data}

We performed a Monte Carlo-type experiment, generating 16 data tables controlling the following factors:
$n$, the number of objects (with levels 120 and 1200); $K$ the number of clusters (levels 3 and 5); $n_k$, the cardinality of the clusters (equal cardinalities and one big cluster with 50\% of the objects and the rest of objects distributed equally); and $\pi$, the probability of the Bernoulli distribution, with levels of well separated clusters ($\pi = 0.1,  0.5,  0.9$ for $K=3$ and $\pi = 0.05, 0.25, 0.5, 0.75, 0.95$ for $K=5$) or fuzzy separated clusters ($\pi = 0.3, 0.5, 0.7$ for $K=3$ and $\pi = 0.2, 0.35, 0.5, 0.65, 0.8$ for $K=5$). 
Table \ref{simulated-tables} presents the characteristics of each of the 16 data tables generated in the experiment, including the factors and the levels.
%Figure \ref{fig:tables} illustrates the first eight tables, where blank cells means zeroes and black ones mean ones.
%\pagebreak

\begin{table}[h]
\centering
\caption{Characteristics of data tables generated, 4 factors with 2 leves each. $n$: number of objects, $K$: number of clusters, $|C_k|$: cardinality of cluster $C_k$, $\pi_k$ probability of membership to $C_k$.}
\scalebox{0.8}{
\begin{tabular}{c|ccccccccccc}
\hline\hline
{Data table} & $n$ & $K$ & {$n_{k}$} & {$\pi$} & {$|C_{1}|$} & {$|C_{2}|,\ldots,|C_{K}|$} & {$\pi_{1}$} & {$\pi_{2}$} & {$\pi_{3}$} & {$\pi_{4}$} & {$\pi_{5}$} \\
\hline
1 & 120  & 3 & = & separated & 40 & 40 & 0,1 & 0.5 & 0.9 &  &  \\
2 & 120  & 3 & = & fuzzy     & 40 & 40 & 0.3 & 0.5 & 0.7 &  &  \\
3 & 120  & 3 & $\neq$ & separated & 60 & 30 & 0.1 & 0.5 & 0.9 &  &  \\
4 & 120  & 3 & $\neq$ & fuzzy & 60 & 30 & 0.3 & 0.5 & 0.7 &  &  \\
5 & 120  & 5 & = & separated & 24 & 24 & 0.05 & 0.25 & 0.5 & 0.75 & 0.95 \\
6 & 120  & 5 & = & fuzzy & 24 & 24 & 0.2 & 0.35 & 0.5 & 0.65 & 0.8 \\
7 & 120  & 5 & $\neq$ & separated & 60 & 15 & 0.05 & 0.25 & 0.5 & 0.75 & 0.95 \\
8 & 120  & 5 & $\neq$ & fuzzy & 60 & 15 & 0.2 & 0.35 & 0.5 & 0.65 & 0.8 \\
9 & 1200  & 3 & = & separated & 40 & 40 & 0,1 & 0.5 & 0.9 &  &  \\
10 & 1200  & 3 & = & fuzzy     & 40 & 40 & 0.3 & 0.5 & 0.7 &  &  \\
11 & 1200  & 3 & $\neq$ & separated & 60 & 30 & 0.1 & 0.5 & 0.9 &  &  \\
12 & 1200  & 3 & $\neq$ & fuzzy & 60 & 30 & 0.3 & 0.5 & 0.7 &  &  \\
13 & 1200  & 5 & = & separated & 24 & 24 & 0.05 & 0.25 & 0.5 & 0.75 & 0.95 \\
14 & 1200  & 5 & = & fuzzy & 24 & 24 & 0.2 & 0.35 & 0.5 & 0.65 & 0.8 \\
15 & 1200  & 5 & $\neq$ & separated & 60 & 15 & 0.05 & 0.25 & 0.5 & 0.75 & 0.95 \\
16 & 1200  & 5 & $\neq$ & fuzzy & 60 & 15 & 0.2 & 0.35 & 0.5 & 0.65 & 0.8 \\
\hline\hline
\end{tabular}
}
\label{simulated-tables}
\end{table}

%\begin{figure}[h!]
%\centering
%\subfigure[Table 1]{\includegraphics[scale=0.200]{Table1}}
%\subfigure[Table 2]{\includegraphics[scale=0.200]{Table2}}
%\subfigure[Table 3]{\includegraphics[scale=0.200]{Table3}}
%\subfigure[Table 4]{\includegraphics[scale=0.200]{Table4}}
%\subfigure[Table 5]{\includegraphics[scale=0.200]{Table5}}
%\subfigure[Table 6]{\includegraphics[scale=0.200]{Table6}}
%\subfigure[Table 7]{\includegraphics[scale=0.200]{Table7}}
%\subfigure[Table 8]{\includegraphics[scale=0.200]{Table8}}
%\caption{Illustration of the tables generated with $n=120$, blank cells mean 0's, black ones mean 1's.}
%\label{fig:tables}
%\end{figure}

%\pagebreak

\section{Results}
\label{sec:results}

We have compared the results obtained with the five metaheuristics (SA, TA, TS, GA, AC) with two classical methods: PAM (partitioning around medoids) when using the $L_1$ dissimilarity index, and hierarchical clustering (HC) using average linkage.
For each heuristic, a parameter calibration was performed, exploring different values for each parameter. After this calibration, parameters selected for this article were:
\begin{itemize}
\item Simulated annealing: $\chi_0 = 0.95$, $L=50$; $\gamma=0.91$, $\epsilon = 0.01$.
\item Threshold accepting: $Th_{0}=100$, $maxiter=50$, $\gamma=0.9$, $\epsilon=0.01$.
\item Tabu search: $maxiter=150$, $|T|=5$, $s=0.1|N(P)|$.
\item Genetic algorithm: $p_m = %p_{m1} \times p_{m2} = 0.5 \times 0.2 = 
0.1$, $p_c=0.8$, $M= 20$, $maxiter = 500$, $\epsilon = 0.01$.
\item Ant colony: $\alpha = 0.5$, $\beta = 0.2$, $\rho = 0.5$, $M=10$, $maxiter = 500$, $\epsilon = 0.01$.
\end{itemize}

In Table \ref{table:results} we report, for a multistart of size 100, the best value of $W$ obtained so far by any method (noted $W^*$), the mean value of $W$ for each method (noted $\xbar{W}$) and the attraction rate $a_r$ or percentage of times that $W^*$ was obtained (up to a relative error of 0.05).

\begin{table}[h]
\caption{Results summary with $\delta_{L_1}$ for a multistart of size 100. $W^*$ is the best value obtained by any method, $a_r$ the attraction rate of $W^*$ for each method, and $\xbar{W}$ the mean value for each method}
\scalebox{0.75}{
\begin{tabular}{p{9mm}|p{7mm}p{7mm}p{7mm}p{5mm}p{7mm}p{5mm}p{7mm}p{5mm}p{8mm}p{5mm}p{8mm}p{7mm}p{7mm}p{7mm}}
\hline\hline
\multirow{2}*{Table} & 
\multirow{2}*{$W^*$} &
\multicolumn{2}{l}{SA} & 
\multicolumn{2}{l}{TA} & 
\multicolumn{2}{l}{TS} & 
\multicolumn{2}{l}{GA} & 
\multicolumn{2}{l}{AC} & 
\multicolumn{2}{l}{PAM} & 
\multicolumn{1}{l}{HC} \\
 & & $a_r$ & $\xbar{W}$ & $a_r$ & $\xbar{W}$ & $a_r$ & $\xbar{W}$ & $a_r$ & $\xbar{W}$ & $a_r$ & $\xbar{W}$ & $a_r$ & $\xbar{W}$ & $\xbar{W}$\\
\hline
1 & 414 & 7\% & 431 & 2\% & 432 & 1\% & 444 & 0\% & 648 & 0\% & 978 & 0\% & 421 & 445\\
2 & 744 & 4\% & 750 & 0\% & 751 & 1\% & 757 & 0\% & 849 & 0\% & 1017 & 0\% & 780 & 790 \\
3 & 387 & 0\% & 412 & 0\% & 412 & 0\% & 412 & 0\% & 605 & 0\% & 901 & 100\% & 387 & 412\\
4 & 367 & 3\% & 387 & 0\% & 429 & 0\% & 430 & 0\% & 611 & 0\% & 901 & 0\% & 400 & 429\\
5 & 424 & 2\% & 456 & 5\% & 444 & 0\% & 473 & 0\% & 688 & 0\% & 951 & 0\% & 426 & 451\\
6 & 587 & 100\% & 587 & 0\% & 607 & 0\% & 620 & 0\% & 797 & 0\% & 963 & 0\% & 600 & 637\\
7 & 293 & 0\% & 326 & 0\% & 325 & 0\% & 345 & 0\% & 576 & 0\% & 762 & 100\% & 293 & 305\\
8 & 513 & 1\% & 525 & 4\% & 522 & 0\% & 559 & 0\% & 720 & 0\% & 853 & 0\% & 542 & 543\\
9 & 4641 & 0\% & 4868 & 0\% & 5439 & 0\% & 4928 & 0\% & 8682 & 0\% & 11350 & 100\% & 4641 & 4983\\
10 & 7775 & 1\% & 7880 & 0\% & 8561 & 0\% & 8210 & 0\% & 11204 & 0\% & 11462 & 0\% & 7841 & 8385\\
11 & 4137 & 100\% & 4137 & 0\% & 4156 & 0\% & 9639 & 0\% & 4161 & 0\% & 9636 & 100\% & 4137 & 4379\\
12 & 4179 & 100\% & 4179 & 0\% & 9582 & 0\% & 9582 & 0\% & 4207 & 0\% & 9681 & 100\% & 4179 & 4494\\
13 & 3003 & 0\% & 4304 & 0\% & 4932 & 0\% & 4523 & 0\% & 11106 & 0\% & 10642 & 100\% & 3003 & 3277\\
14 & 6549 & 0\% & 7218 & 0\% & 7364 & 0\% & 7272 & 0\% & 11053 & 0\% & 11288 & 100\% & 6549 & 7192\\
15 & 3165 & 10\% & 4512 & 1\% & 8114 & 5\% & 7985 & 0\% & 3201 & 0\% & 7883 & 100\% & 3165 & 3337\\
16 & 5812 & 10\% & 6114 & 10\% & 6442 & 5\% & 10558 & 0\% & 5896 & 0\% & 9369 & 100\% & 5812 & 6270\\
\hline\hline
\end{tabular}
}
\label{table:results}
\end{table}

\pagebreak

\section{Concluding remarks}
\label{sec:conclusion}

Generally speaking, with simulated annealing we obtain good results, although PAM obtains good results in some cases. Threshold accepting sometimes reaches the optimum and tabu search only in two cases. Population based heuristics did not get good results with our implementation. It is worth noting that hierarchical clustering never obtained the optimum. Even if we do not report running times, SA is fast enough to be competitive. The main drawback of using heuristics, is tuning the parameters, though SA may have a standard choice.

\subsection*{Acknowledgements}
This research was partially supported by the University of Costa Rica (CIMPA project 821-B1-122 and the Graduate Program in Mathematics) and the Costa Rica Institute of Technology. The supports are gratefully acknowledged.


\begin{thebibliography}{99}%
%%%%%%%%%%%%%%%%%%%%%%%%%%%%%%%%%%

\bibitem{Aarts} % book
Aarts, E.; Korst, J.:
Simulated Annealing and Boltzmann Machines.
John Wiley \& Sons, Chichester (1990)

\bibitem{Borkulo} % article
Borkulo, C.D. van, Borsboom, D., Epskamp, S., Blanken, T.F., Boschloo, L., Schoevers, R.A., Waldorp, L.J.:
A new method for constructing networks from binary data.
Scient. Rep. \textbf{4} (2014) doi: 10.1038/srep05918

\bibitem{Dorigo} % book
Bonabeau, E., Dorigo, M., Therauluz, G.:
Swarm Intelligence. From Natural to Artificial Systems. 
Oxford University Press, New York (1999)

\bibitem{Demey} % article
Demey, J.R., Vicente-Villardón, J.L., Galindo-Villardón, M.P., Zambrano, A.Y.:  
Identifying molecular markers associated with classification of genotypes by external logistic biplots. 
Bioinform. \textbf{24}(24), 2832--2838 (2008)

\bibitem{Dueck:Scheuer} % article
Dueck, G.; Scheuer, T.: 
Threshold accepting: A general purpose optimization algorithm appearing superior to simulated annealing, 
J. Comput. Phys. \textbf{90}(1), 161--175 (1990)

\bibitem{Everitt} % book
Everitt, B.S.: Cluster Analysis. Edward Arnold, London (1993)

\bibitem{Glover} % article
Glover, F.: Tabu search – Part I. ORSA J. on Comput., \textbf{1}, 190--206 (1989).

\bibitem{Goldberg} % book
Goldberg, D.E.: Genetic Algorithms in Search Optimization and Machine Learning. 
Addison-Wesley, Reading (1989)

\bibitem{Jajuga:1987} % article
Jajuga, K.:
A clustering method based on the $L_1$-norm.
Comput. Stat. \& Data Analysis \textbf{5}(4), 357--371 (1987)
doi: 10.1016/0167-9473(87)90058-2

\bibitem{Jeliazkov} % contribution
Jeliazkov, I., Rahman, M.A.:
Binary and ordinal data analysis in Economics: Modeling and estimation. 
In: Yang, X.S. (ed.) Mathematical Modeling with Multidisciplinary Applications, pp. 1--31.
John Wiley \& Sons, New York (2012)

\bibitem{Kaufman:Roosseuw}
Kaufman, L., Roosseuw, P.:
Finding Groups in Data. An Introduction to Cluster Analysis.
John Wiley \& Sons, New York (2005)

\bibitem{Kirkpatrick} % article
Kirkpatrick, S., Gelatt, D., Vecchi, M.P.: Optimization by simulated annealing. 
Science, \textbf{220}, 671--680 (1983)

\bibitem{Koln} % contribution
Piza, E., Trejos, J., Murillo, A.:  
Clustering with non-Euclidean distances using combinatorial optimisation techniques. 
In: Blasius, J., Hox, J., de Leeuw, E., Schmidt, P. (eds.)
Social Science Methodology in the New Millennium,% -- Proceedings of the Fifth International Conference on Logic and Methodology, 
paper number P090504.
Leske + Budrich, Darmstadt (2002) %ISBN 978-3810033161

\bibitem{Salas} % article
Salas-Eljatiba, C., Fuentes-Ramirez, A., Gregoire, T.G., Altamirano, A., Yaitula, V.:
A study on the effects of unbalanced data when fitting logistic regression models in ecology.
Ecological Indicators, \textbf{85}, 502--508 (2018)

\bibitem{Spath} % book
Sp\"ath, H.:
Cluster Dissection and Analysis. Theory, Fortran programs, Examples.
Ellis Horwood, Chichester (1985)

\bibitem{IFCS:1998} % contribution
Trejos, J.; Murillo, A.; Piza, E.:
Global stochastic optimization techniques applied to partitioning.
In: Rizzi, A., Vichi, M., Bock, H.-H. (eds.)
Advances in Data Science and Classification, pp. 185--190.
Springer, Berlin (1998)

\bibitem{IEEE:2014} % contribution
Trejos, J., Villalobos, M., Murillo, A., Chavarr\'ia, J., Fallas, J.J.: 
Evaluation of optimization metaheuristics in clustering. 
In: Travieso, C.M., Arroyo, J., Ram\'irez, M. (eds.)
IEEE International Work-Conference on Bioinspired Intelligence, pp. 154--161. 
IEEE, Liberia (2014) doi: \url{10.1109/IWOBI.2014.6913956}

\bibitem{Zhang} % book
Zhang, H., Singer, B.:
Recursive Partitioning in the Health Sciences.
Springer, New York (1999)

\end{thebibliography}
\end{document}